\documentclass{article}
\usepackage[latin9]{inputenc}
\usepackage{amsmath}
\usepackage{amssymb}
\usepackage{graphicx}

\makeatletter

\providecommand{\tabularnewline}{\\}

\usepackage{spconf}

\title{effect of Super Resolution on High dimensional features for Unsupervised face recognition in the wild}
\name{Ahmed ElSayed, Ausif Mahmood, Tarek Sobh}
\address{Department of Computer Science and Engineering, University of Bridgeport, Bridgeport, CT 06604, USA\\
Emails: aelsayed@my.bridgeport.edu, \{mahmood,sobh\}@bridgeport.edu}

\let\textquotedbl="
\makeatother

\begin{document}
\maketitle
\begin{abstract}
Majority of the face recognition algorithms use query faces captured
from uncontrolled, in the wild, environment. Often caused by the cameras\textquoteright{}
limited capabilities, it is common for these captured facial images
to be blurred or low resolution. Super resolution algorithms are therefore
crucial in improving the resolution of such images especially when
the image size is small requiring enlargement. This paper aims to
demonstrate the effect of one of the state-of-the-art algorithms in
the field of image super resolution. To demonstrate the functionality
of the algorithm, various before and after 3D face alignment cases
are provided using the images from the Labeled Faces in the Wild (lfw).
Resulting images are subject to testing on a closed set face recognition
protocol using unsupervised algorithms with high dimension extracted
features. The inclusion of super resolution algorithm resulted in
significant improved recognition rate over recently reported results
obtained from unsupervised algorithms. 
\end{abstract}

\section{Introduction}

\label{sec:intro}

Majority of the surveillance cameras are installed outdoors and therefore,
the captured images are likely to be impacted by the surrounded environment.
These images are called \textquotedblleft images in the wild\textquotedblright{}
and when used for face recognition, their size and resolution affect
the accuracy of facial recognition. Current literature offers limited
studies focusing on this problem. Existing studies\cite{SR_video_2007,SR_video_2007_2,SR_video_2012,SR_video_2013}
mostly focus on a video or a multi-frame based super resolution construction
of the low resolution face images. In these, the authors focus on
performance of traditional face recognition techniques on lower and
super resolution faces constructed from multi-frame videos. In real
world applications however, the problem at hand often has a single
query image and not multi-frame video. 

Other relevant studies \cite{SR_face_recognition_2012,SRCNN_faces_2016}
utilize single image super-resolution algorithms to study the performance
of face recognition algorithms on varying face resolutions. However,
these studies did not investigate the performance of face recognition
using high dimension features. Furthermore, both studies utilized
test datasets which include images captures in controlled environments. 

This research studies the performance of unsupervised face recognition
for labeled faces in the wild (lfw) dataset \cite{LFWTech,LFWTechUpdate}
using a single image super-resolution algorithm. The effect of the
algorithm on high dimensional features used in the face recognition
process is investigated. Each image in the dataset is 3D aligned and
frontalized using face frontalization algorithm as proposed in \cite{HHPE:CVPR15:frontalize}. 

The main contribution of this paper is: 
\begin{itemize}
\item Applying Local Binary Pattern (LBP) and Multi-Scale LBP features on
captured faces in the wild and using calculated features in unsupervised
closed set face recognition. 
\item Studying the effect of single image super-resolution algorithm vs
bicubic scaling on unsupervised face recognition in the wild. 
\item Examining the order of applying face frontalization and image sharpness
(super-resolution) process. 
\end{itemize}
Following sections include details of the super-resolution algorithm
and the discussion regarding the LBP high dimension features. After
the comparative analysis a description of the proposed experiment
and the techniques utilized are provided. This is followed by the
explanation of the algorithm results. Lastly, conclusions and discussions
are given in the final section. 

\section{Single Image Super-Resolution}

\begin{figure}[htb]
\noindent\begin{minipage}[b]{1\linewidth}%
\centering \centerline{\includegraphics[width=8.5cm]{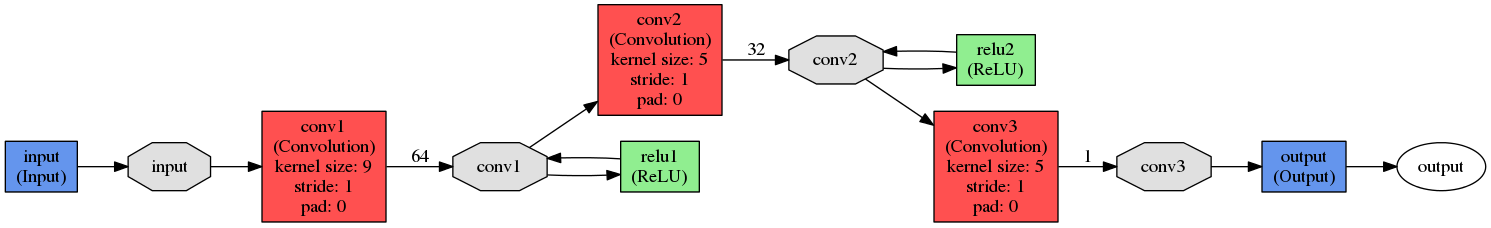}} %
\end{minipage}

\caption{Super-Resolution using Convolutional Neural Network (SRCNN) algorithm
used in the paper}

\label{fig:res-1-1}
\end{figure}

Super-Resolution algorithm is used to enhance image resolution and
to provide additional details of the input image. In this work, a
super-resolution image algorithm based on Convolutional Neural Network
(CNN) is used as also described in \cite{SRCNN}. The system first
generates low resolution higher dimension image from the input image
using bicubic interpolation. This image is then applied to a CNN network
structure as shown in Figure \ref{fig:res-1-1} to improve the image
peak signal to noise ratio (PSNR) for generating a higher resolution
image that should be close to the original image in quality. The utilization
of CNN makes the proposed algorithm superior to other similar SR techniques
that generate mapping from low to high resolution images due to its
simplicity and the resulting higher PSNR compared to other approaches. 

\section{High Dimensional Features }

Unsupervised face recognition found a recent interest due to the capability
of handling unlabeled faces, especially in closed datasets as in \cite{LBP_face,SURF-Face,ElSayed2015}.
The research on high dimensional features has provided remarkable
results in face recognition and verification, particularly with supervised
learning as in \cite{HIGHDIMLBP_2013,lfw_protocol}. These features
however have not been sufficiently explored using unsupervised techniques.
This section demonstrates the utilization of one of those features
using unsupervised metric for closed set protocol on the lfw dataset. 

In \cite{LBP_face} LBP features have provided remarkable unsupervised
face recognition outcomes for faces in controlled environment. Therefore,
the same Chi square metric given in equation \ref{eq:1} is used in
the testing of the extracted features from the lfw dataset. 

\begin{equation}
\chi^{2}\left(X,Y\right)=\sum_{i,,j}\frac{\left(x_{i,j}-y_{i,j}\right)^{2}}{x_{i,j}+y_{i,j}},\label{eq:1}
\end{equation}

where,$X$ and $Y$ are the histograms to be compared, $i$ and $j$
are the indices of the $i$-th bin in histogram corresponding to the
$j$-th local region. 

In this test, three types of LBP features are demonstrated. The first
one is the regular uniform LBP features extracted from frontalized
faces by dividing the 90x90 face into 10x10 blocks, each being 9x9
pixels. Following this (8,2) ($LBP_{8,2}^{u2}$) neighborhoods are
calculated for each block as in \cite{LBP_face}, The histograms of
all blocks are then concatenated together to form a single vector
representation for the face image to be used in equation \ref{eq:1}.
The output vector of this calculation will be 5900 in length.

\begin{figure}[htb]
\noindent\begin{minipage}[b]{1\linewidth}%
 \centering \centerline{\includegraphics[width=7cm]{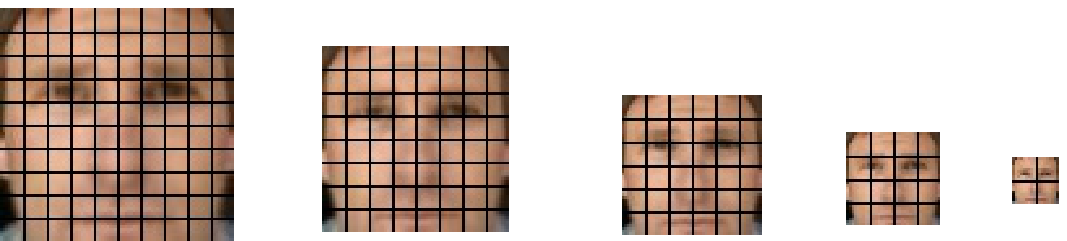}}
\centerline{(a)}%
\end{minipage}

\noindent\begin{minipage}[b]{1\linewidth}%
 \centering \centerline{\includegraphics[width=7cm]{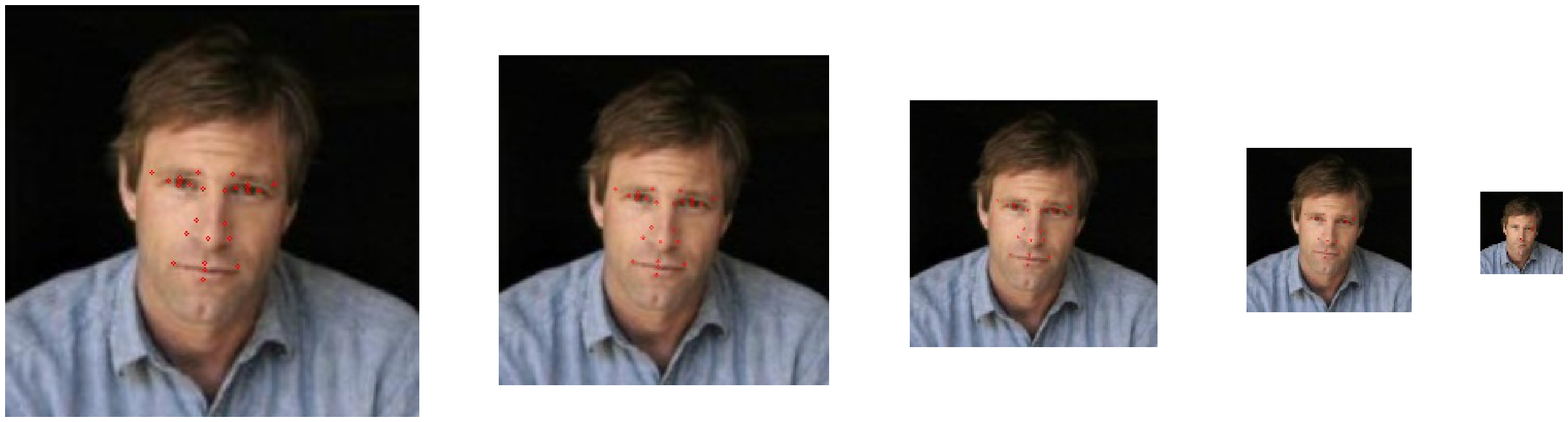}}
\centerline{(b)}%
\end{minipage}\caption{Two high dimensional LBP features a)Multi-Scale LBP b)HighDimLBP \cite{HIGHDIMLBP_2013}.}

\label{fig:res-1} 
\end{figure}

The second type of LBP is a Multi-Scale representation. The frontalized
face is scaled down 5 times, and for each scale the image is divided
to 10x10 blocks 9x9 pixels each as shown in Figure \ref{fig:res-1}
a. The $LBP_{8,2}^{u2}$ histogram is then calculated again for each
block at each scale and all histograms are concatenated together to
form a vector representation for the face with a length of 12980. 

The final LBP type is the HighDimLBP introduced in \cite{HIGHDIMLBP_2013},
where the faces are not frontalized but instead an accurate landmarks
detection technique is used to obtain facial landmarks. Then for each
landmark in the 300x300 image a grid of 40x40 centered at each landmark
point is constructed and $LBP_{8,2}^{u2}$ is calculated over each
10x10 pixels block as shown in Figure \ref{fig:res-1} b. Following
this, all histograms from all blocks for all landmark points on the
5 different scales are concatenated together to form a vector representation
of the face image. The length of this vector for one image is 127440which
is significantly long and computationally expensive. Therefore, in
some cases, the size is reduced to 400 using the principle component
analysis (PCA) to improve the computational performance. Similar approach
has also been used used in \cite{lfw_protocol,HIGHDIMLBP_2013}. 

A comparison is made between these three types to obtain the best
technique in the proposed experiments. The next section details the
experiment results. 

\section{EXPERIMENT DESCRIPTION}

\label{sec:illust}

\begin{figure}[htb]
\noindent\begin{minipage}[b]{1\linewidth}%
 \centering \centerline{\includegraphics[width=7cm]{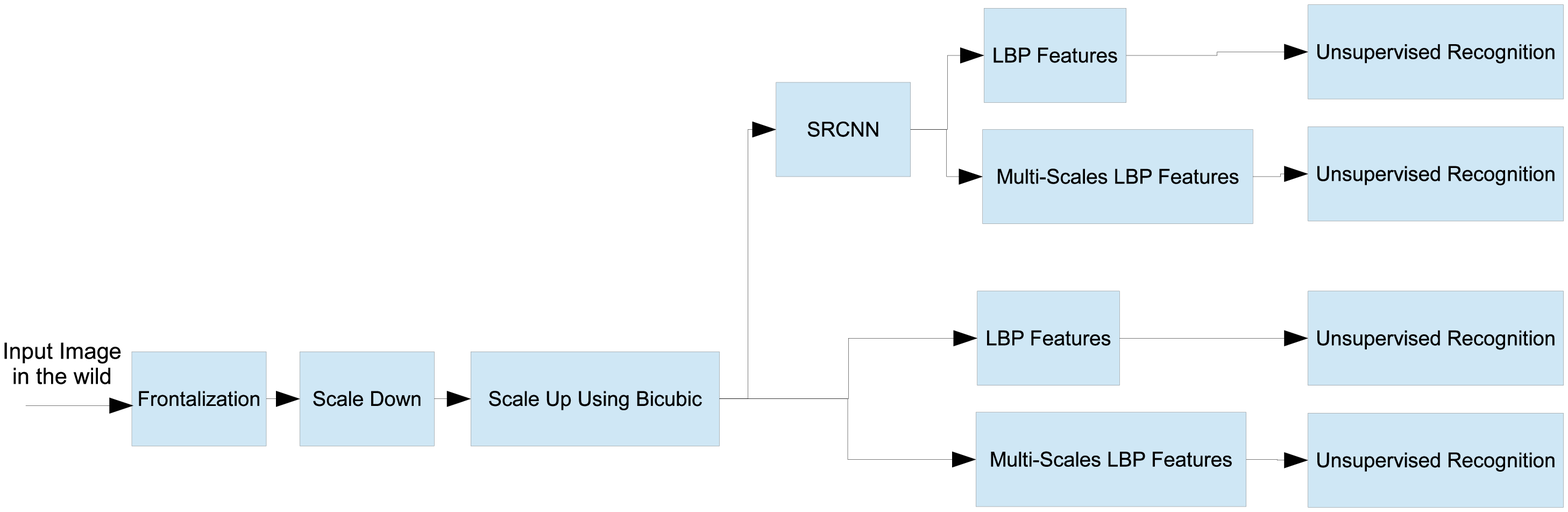}}
\centerline{(a)}%
\end{minipage}

\noindent\begin{minipage}[b]{1\linewidth}%
 \centering \centerline{\includegraphics[width=7cm]{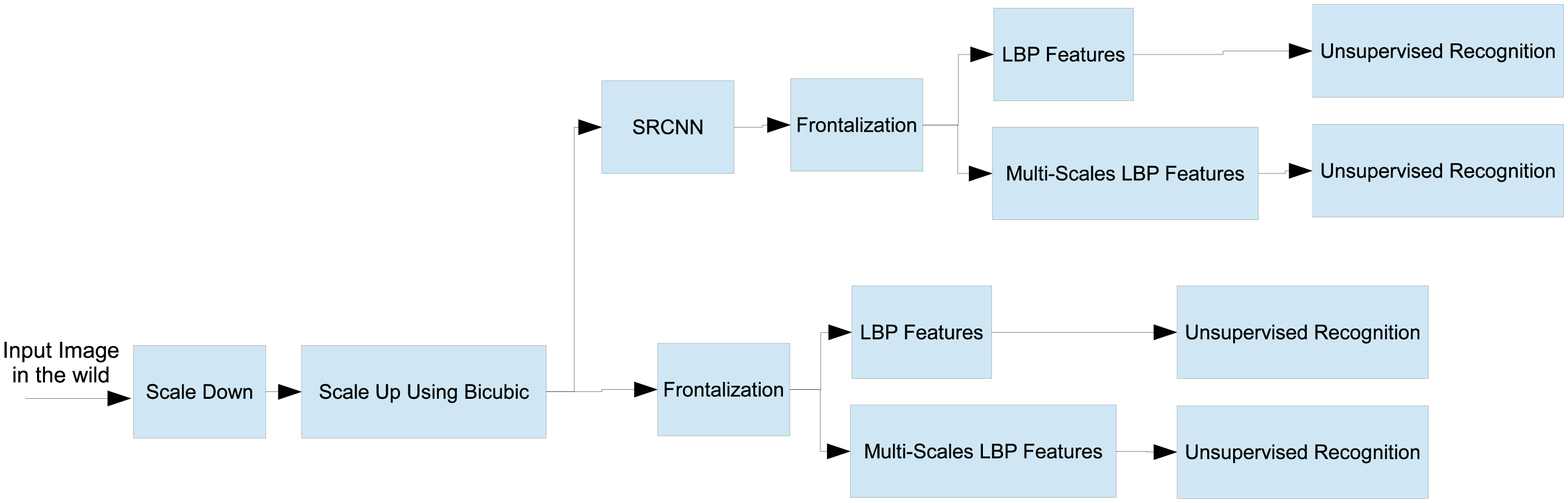}}
\centerline{(b)}%
\end{minipage}\caption{Proposed experiments a)Frontalization first b)Scaling and SR first.}

\label{fig:res-1-2-1} 
\end{figure}

This paper proposes two experiments to examine the effect of image
super-resolution and the order of applying it with frontalization
to unsupervised face recognition process based on the features described
in section 3. These two experiments are detailed in the following: 
\begin{enumerate}
\item Apply face frontalization first. The work flow of this experiment
is depicted in Figure \ref{fig:res-1-2-1} a, and can described as
in the following:
\begin{enumerate}
\item Detect and frontalize face from the original sized image (250x250
in this case).
\item Scale down the face image by scale of 3, To assuming the case of detecting
face at that size (if the frontalized face image size is 90x90 the
resulting image size will be 30x30, an appropriate size for face detection
techniques\footnote{Minimum detection size of Haar Cascade classifier for face detection
is 24x24}).
\item Scale up the face image again by scale of 3 using bicubic technique.
\item Apply the SRCNN algorithm to the scaled face to generate a super-resolution
version.
\item Extract uniform local binary pattern (LBP) features from the SR-image
by dividing it into 10x10 blocks and concatenating the histograms
of all blocks together. This step is applied on both bicubic and super-resolution
scaled faces to compare the performance of the recognition process.
\item For Multi-Scale LBP the face image is scaled down for five scales
as shown in Figure \ref{fig:res-1} a. The histograms of all blocks
and scales are concatenated together. This step will be reapplied
on both bicubic and super-resolution scaled faces to compare the performance
of the recognition process.
\item Calculate $\chi^{2}$distances between the extracted features to obtain
the minimum distances between the query images and the prob ones using
equation 1.
\end{enumerate}
\item Process face images prior to frontalization. The work flow of this
experiment is as shown in Figure \ref{fig:res-1-2-1} b, and can described
in the following steps::
\begin{enumerate}
\item Scale down the face image by scale of 3.
\item Scale up the face image again by scale of 3 using bicubic technique.
\item Apply the SRCNN algorithm to the scaled image to generate a super-resolution
version.
\item Extract frontalized faces from both bicubic images and super-resolution
ones for performance comparison.
\item Calculate features and distances as described in steps e to g in experiment
1.
\end{enumerate}
\end{enumerate}

\section{RESULTS}

\label{sec:foot}

\begin{figure}[htb]
\noindent\begin{minipage}[b]{1\linewidth}%
\centering \centerline{\includegraphics[width=6cm]{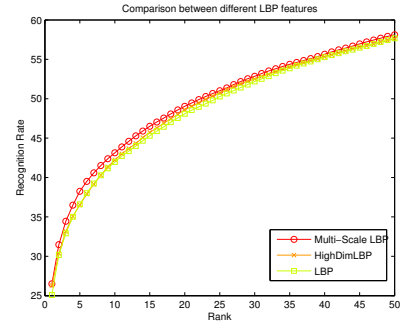}} %
\end{minipage}

\caption{Average percentage recognition rate for 3 different LBP features.}

\label{fig:res-1-1-1}
\end{figure}

\begin{table}
\centering \centerline{%
\begin{tabular}{|c|c|}
\hline 
 & Rank 1 rate (\%)\tabularnewline
\hline 
\hline 
LBP & 25.09\tabularnewline
\hline 
Multi-Scale LBP & 26.51\tabularnewline
\hline 
HighDimLBP & 26.30\tabularnewline
\hline 
HighDimLBP+PCA \cite{lfw_protocol} & 16.50\tabularnewline
\hline 
\end{tabular}}

\caption{Average rank 1 recognition rate using different features. }

\label{table-1-1}
\end{table}

The proposed comparison and experiments have been tested on the Labeled
Faces in the Wild (lfw) dataset \cite{LFWTech,LFWTechUpdate} using
closed set face recognition protocol proposed in \cite{lfw_protocol}.
In this protocol, 10 groups are extracted from the entire dataset,
each group having two sets; gallery and genuine prob. Both the gallery
and prob sets included images of 1000 different persons. Each gallery
set contains 1000 images, one image per person, with the size of the
prob set varying from one group to another with an average of 4500
images for the same 1000 persons in the gallery set. The recognition
rates calculated in this paper represent the average recognition rates
over the all 10 protocol groups. 

In this work the faces are detected using Histograms of oriented Gradients
(HoG) algorithm proposed in \cite{Face_detection_HOG}using python.
For each detected face, an algorithm for landmarks detection based
on regression tree is then used for face landmarks detection as in
\cite{Face_landmarks_2014} using python\footnote{Python wrapper for dlib and OpenCV libraries}.
Experiment 2 included some cases where the HoG based face detection
algorithm failed to detected faces due to the effect of image scaling.
Therefore, an alternative backup face detection algorithm which is
based on Adaboost Haar Cascade \cite{viola_jones_2001,AbualkibashEM13}
is used in cases where no faces were detected in the image\footnote{Adaboost Haar Cascade classifier is known to have higher face detection
rate but with large number of false positives \cite{AbualkibashEM13,viola_jones_2001}}.

First, a comparison between the three different types of LBP features
has been applied to this dataset and Chi square metric has been used
as an unsupervised face recognition metric. As shown in figure \ref{fig:res-1-1-1}
the Multi-Scale LBP features outperform other LBP types, especially
the method of using HighDimLBP+PCA listed in \cite{lfw_protocol}.
However, as shown in table \ref{table-1-1} both Multi-Scale LBP and
HighDimLBP with Chi square distance have close recognition rates.
It should also be noted that the computation time of Chi square distance
for HighDimLBP is significantly high compared to other LBP types due
to the length of the vector representation.

For the two experiments, the super-resolution based on convolutional
neural network (SRCNN) algorithm is implemented using Caffe library
and tested using Matlab. But, instead of applying SR algorithm on
the y component only of the ycbcr domain (because it is the one with
the high frequencies), in this test the SR algorithm is applied on
the three channels of the RGB domain to enhance both the edges and
colors of the estimated pixels by the bicubic scaling.

For the protocol used, the faces are first frontalized as in \cite{HHPE:CVPR15:frontalize}
and an unsupervised face recognition based on LBP and Multi-Scale
LBP features is utilized to create a baseline for comparison. The
results of proposed experiments are marked as lfw3D in the tables
and figures. The results of experiment 1 of the bicubic scaling is
marked as lfw3D bicubic 3 channels where as for the super-resolution
version they are called lfw3D SR 3 channels. Experiment 2 results
of bicubic scaling are marked as lfw bicubic 3 channels original cropped
where as the super-resolution version is marked as lfw SR 3 channels
original cropped.

\begin{figure}[htb]
\noindent\begin{minipage}[b]{1\linewidth}%
 \centering \centerline{\includegraphics[width=6cm]{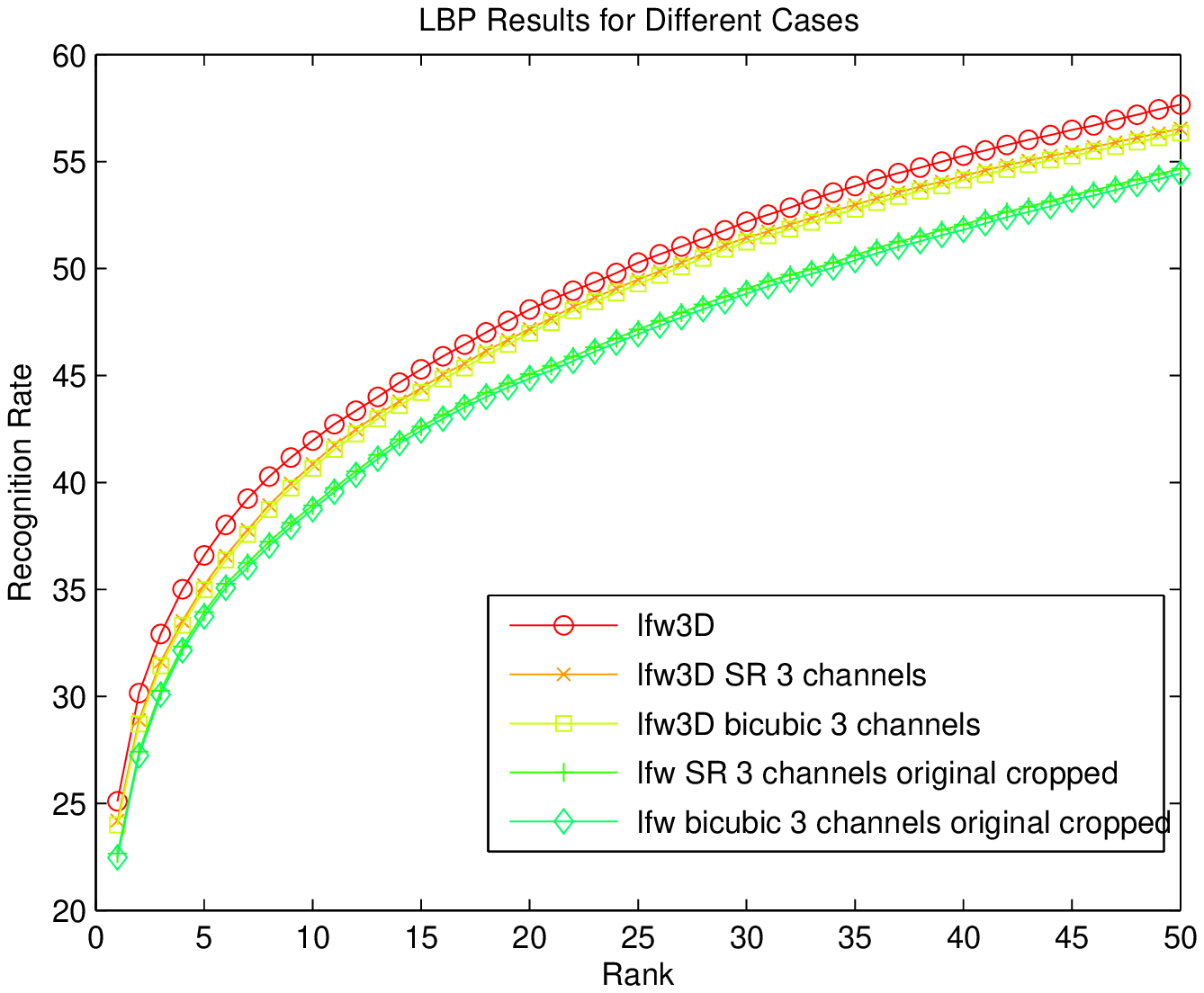}}
\centerline{(a)}%
\end{minipage}

\noindent\begin{minipage}[b]{1\linewidth}%
 \centering \centerline{\includegraphics[width=6cm]{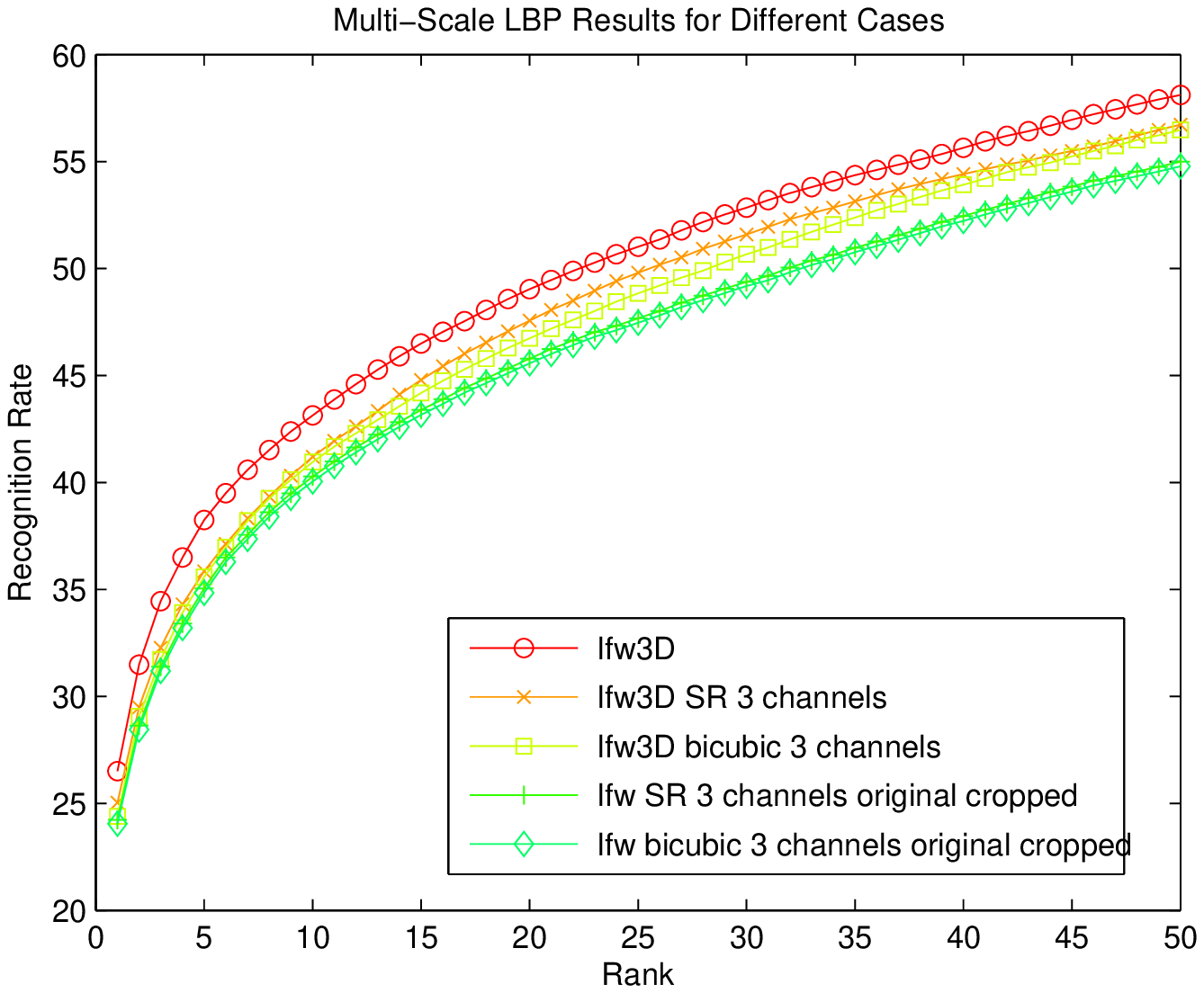}}
\centerline{(b)}%
\end{minipage}\caption{Average percentage recognition rate results for both a)LBP b)Multi-scale
LBP.}

\label{fig:res-1-2} 
\end{figure}

As shown in figure \ref{fig:res-1-2}, the super-resolution algorithm
enhances the recognition rates for both LBP and Multi-Scale LBP features
over bicubic scaled version in both experiments. However, both are
still lower than the baseline recognition rate. Moreover, the recognition
rate of experiment 1 is superior to the one collected from experiment
2. This is significant since it indicates that applying face frontalization
prior to scaling and sharpening process provides better results than
scaling all the images up and frontalizing the detected face. It can
also be observed that Multi-Scale LBP performs better in both experiments
and outperforms all other features used in the presented unsupervised
test. 



\begin{table}
\centering \centerline{%
\begin{tabular}{|c|c|c|}
\hline 
 & LBP & Multi-Scale LBP\tabularnewline
\hline 
\hline 
lfw3D & 25.09 & 26.51\tabularnewline
\hline 
lfw3D SR 3 channels & 24.09 & 24.93\tabularnewline
\hline 
lfw3D bicubic 3 channels & 23.99 & 24.38\tabularnewline
\hline 
lfw SR 3 channels orig. & 22.65 & 24.22\tabularnewline
\hline 
lfw bicubic 3 channels orig. & 22.46 & 24.04\tabularnewline
\hline 
\end{tabular}}

\caption{Average rank 1 recognition rate of all cases in the experiments.}

\label{table-1}
\end{table}

\section{CONCLUSION}

\label{sec:copyright}

This work utilized an unsupervised face recognition with images from
the Labeled Faces in the Wild (lfw) dataset with LBP and Multi-Scale
LBP based extracted features. The results indicate that Multi-Scale
LBP outperforms both LBP and HighDimLBP features with reasonable extraction
and distance calculation time. Two experiments have also been introduced
to measure the performance of applying single image super-resolution
algorithm on faces captured in the wild and the effect of order of
applying it with face frontalization algorithm. It can be concluded
that applying super resolution on frontalized faces provides better
results as opposed to applying super resolution first. This is because
face frontalization uses interpolation to calculate some pixel values,
similar to bicubic scaling, which will get enhanced with super-resolution
techniques. The results also indicate that applying super-resolution
on bicubic scaled faces shows slight enhancement in unsupervised face
recognition process for both experiments with the two types of features. 

\bibliographystyle{IEEEbib}
\bibliography{refrences}

\end{document}